%% file: arXiv.tex
\documentclass[10pt,twocolumn,letterpaper]{article}

\usepackage[pagenumbers]{cvpr} 


\definecolor{cvprblue}{rgb}{0.21,0.49,0.74}
\usepackage[pagebackref,breaklinks,colorlinks,allcolors=cvprblue]{hyperref}
\usepackage{multirow}

\def\paperID{8762} 
\def\confName{ICCV}
\def\confYear{2025}

\title{Large Language Model Guided Progressive Feature Alignment for Multimodal UAV Object Detection}

\author{
  Wentao Wu${^1}$, Chenglong Li${^1}$, Xiao Wang${^2}$, Bin Luo${^2}$, Qi Liu${^3}$ \\
  ${^1}${School of Artificial Intelligence, Anhui University, Hefei, China} \\
  ${^2}${School of Computer Science and Technology, Anhui University, Hefei, China} \\
  ${^3}${University of Science and Technology of China, Hefei, China} \\
  \textit{wuwentao0708@163.com, lcl1314@foxmail.com, 
  \{xiaowang, luobin\}@ahu.edu.cn}, qiliuql@ustc.edu.cn
}

\begin{document}
\maketitle
\input{sec_arxiv/0_abstract}    
\input{sec_arxiv/1_intro}

\input{sec_arxiv/2_related}

\input{sec_arxiv/3_meth}
\input{sec_arxiv/4_exper}
\input{sec_arxiv/5_con}

{
    \small
    \bibliographystyle{ieeenat_fullname}
    \bibliography{main}
}


\end{document}

%% file: sec_arxiv/0_abstract.tex
\begin{abstract}

Existing multimodal UAV object detection methods often overlook the impact of semantic gaps between modalities, which makes it difficult to achieve accurate semantic and spatial alignments, limiting detection performance. To address this problem, we propose a Large Language Model (LLM) guided Progressive feature Alignment Network called LPANet, which leverages the semantic features extracted from a large language model to guide the progressive semantic and spatial alignment between modalities for multimodal UAV object detection. To employ the powerful semantic representation of LLM, we generate the fine-grained text descriptions of each object category by ChatGPT and then extract the semantic features using the large language model MPNet. Based on the semantic features, we guide the semantic and spatial alignments in a progressive manner as follows. First, we design the Semantic Alignment Module (SAM) to pull the semantic features and multimodal visual features of each object closer, alleviating the semantic differences of objects between modalities. Second, we design the Explicit Spatial alignment Module (ESM) by integrating the semantic relations into the estimation of feature-level offsets, alleviating the coarse spatial misalignment between modalities. Finally, we design the Implicit Spatial alignment Module (ISM), which leverages the cross-modal correlations to aggregate key features from neighboring regions to achieve implicit spatial alignment. Comprehensive experiments on two public multimodal UAV object detection datasets demonstrate that our approach outperforms state-of-the-art multimodal UAV object detectors. 


\end{abstract}

%% file: sec_arxiv/1_intro.tex
\begin{figure}
\centering
\includegraphics[width=3.3in]{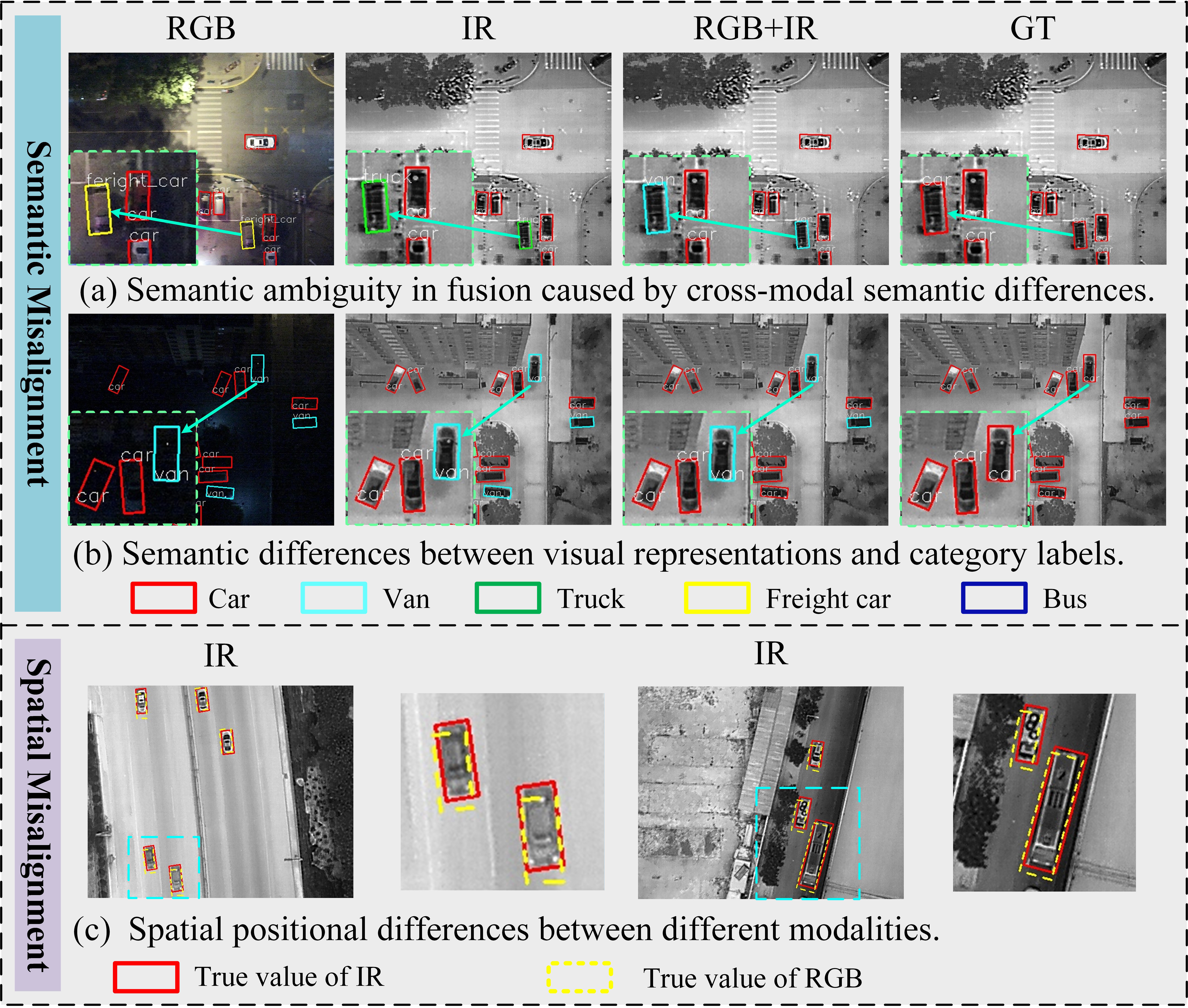}
\caption{Illustration of semantic gap and spatial misalignment issues in multimodal object detection. (a) The semantic gap between different visual modalities causes confusion in the fused region's semantic features, leading to incorrect object category predictions. (b) The semantic gap between visual representations and category labels leads to incorrect object category predictions. (c) Spatial position discrepancies between different visual modalities impact modality fusion.} 
\label{firstimage}
\end{figure}

\section{Introduction} \label{sec:intro}

Multimodal UAV object detection aims to achieve accurate recognition and localization of objects using data from different sensors. Since visible light images provide rich color and detailed information under normal lighting conditions, while infrared images capture abundant object information even in low-light conditions, the combination of visible light and infrared images has attracted widespread attention~\cite{sun2022drone, zhang2021weakly, yuan2024c2former,chen2024weakly}. However, due to the differences in imaging principles, there is usually semantic and spatial misalignment between visible and infrared images, which limits detection performance. 

In multimodal UAV object detection, the first challenge is the semantic misalignment between modalities. Existing methods, such as cross-modal attention mechanisms~\cite{xie2023cross}, shared spatial feature alignment~\cite{yuan2024improving}, and aligning cross-modal contextual information~\cite{he2023multispectral}, are effectively mitigating the semantic misalignment between modalities. However, relying solely on intra-modal information is not enough to fully resolve this issue.
As shown in Figure~\ref{firstimage} (a), different modalities assign completely opposite semantic categories to the same object, causing confusion of semantic information during fusion and ultimately leading to errors in predicting object categories. As shown in Figure~\ref{firstimage} (b), although both the visible and infrared modalities assign the same semantic category to the same object, the semantic gap between the visual modality and the category label results in incorrect semantic category assignment, leading to errors in object category prediction and ultimately affecting the model's detection performance. 

The second challenge is the spatial misalignment between modalities. 
Researchers typically perform preregistration on RGB-IR image pairs, often employing geometric transformation methods such as image cropping~\cite{wang2019deformable}, affine transformation~\cite{xu2021emfusion}, and perspective transformation~\cite{qin2019unsupervised}. However, as shown in Figure~\ref{firstimage} (c), geometric transformations alone are insufficient to resolve spatial misalignment caused by inherent differences in multimodal sensors, as well as issues like imaging time delays. Some existing methods~\cite{zhang2021weakly, yuan2022translation} model differences at the image or feature level to predict spatial offsets, thereby aligning the spatial positions across modalities. However, they overlook the fact that these differences arise not only from spatial offsets but also from semantic and sensor-specific differences, which lead to inaccurate offset estimation and result in imprecise spatial alignment between modalities, limiting the performance of detection.

To address the above two challenges, we consider leveraging the powerful semantic representation capabilities of large language models (LLM) to guide semantic alignment between modalities, eliminating the interference of semantic differences on spatial alignment, and progressively achieving cross-modal feature spatial alignment in a coarse-to-fine manner. Specifically, we utilize ChatGPT to generate fine-grained textual descriptions for each object category and extract semantic features using the large language model MPNet~\cite{song2020mpnet}. Based on these semantic features, we guide semantic and spatial alignment in a progressive manner. 
First, the inherent semantic differences between modalities can lead to semantic ambiguity in the fused features. To address this issue, we design a Semantic Alignment Module (SAM), that maximizes the similarity between the semantic features of each object and the multimodal visual features, bringing them closer in the shared space, thereby alleviating semantic differences between modalities. 
%
Second, due to differences in the acquisition devices of different modalities, there is a spatial misalignment phenomenon in the target positions. To address this issue, we introduce an Explicit Spatial alignment Module (ESM), which enhances object regions in multimodal visual features using the similarity scores calculated in SAM. By integrating semantic relationships into feature-level offset estimation, we achieve coarse spatial alignment between modalities. 
Finally, to further align the features after the coarse alignment by ESM, we design an Implicit Spatial alignment Module (ISM), which aggregates key features from adjacent regions using cross-modal correlations and applies a symmetric consistency loss constraint to achieve implicit spatial alignment of multimodal features. We conduct extensive experiments on the multimodal UAV object detection benchmark datasets, DroneVehicle and VEDAI, fully demonstrating the effectiveness of our proposed framework.

To sum up, we draw the main contributions of this paper as the following four aspects: 

 $\bullet$ We propose a Large Language Model guided Progressive feature Alignment framework called LPANet. By introducing LLM to construct and encode category text descriptions, we achieve semantic feature-guided progressive semantic and spatial alignment. 

 $\bullet$ To alleviate the semantic differences between multimodal features, we design a Semantic Alignment Module that brings the multimodal visual features of the object and the semantic features extracted by LLM closer in the shared space, achieving cross-modal semantic alignment.

 $\bullet$ To address the issue of spatial misalignment between modalities, we design an Explicit Spatial Alignment Module, which incorporates semantic relationships into the estimation of feature-level offsets to achieve coarse spatial alignment across modalities.
 
 $\bullet$ To further align the spatial information based on the coarse alignment by ESM, we design an Implicit Spatial Alignment Module, which aggregates key features from neighboring regions using cross-modal correlations and achieves implicit spatial alignment of the features.

%% file: sec_arxiv/2_related.tex
\section{Related Works}
\label{sec:related}

\subsection{Multimodal Object Detection} 

Early Object detection algorithms~\cite{farhadi2018yolov3, girshick2015region, he2017mask, carion2020end} predominantly used RGB images as input. However, under low-light conditions, relying solely on RGB images makes it difficult to achieve better performance.
Since infrared imaging is not affected by lighting conditions, some researchers have attempted to enhance the accuracy and robustness of object detection algorithms by incorporating infrared data.  
Liu et al.~\cite{liu1611multispectral} proposed using a dual-branch Faster R-CNN~\cite{ren2016faster} framework to separately process visible and infrared images, performing feature fusion at the intermediate convolutional layers, which significantly improved detection accuracy. 
Li et al.~\cite{li2019illumination} integrated a lighting-aware module into the pedestrian detection framework, adaptively fusing multispectral features based on the lighting values of the input images. 
Zhou et al.~\cite{zhou2020improving} proposed the Modal Balance Network (MBNet) to address the issue of modality imbalance during the modality fusion process. Sun et al.~\cite{sun2022drone} designed an uncertainty-aware module to suppress redundant information introduced during modality fusion, quantifying the uncertainty weights of each modality based on lighting values. Zhang et al.~\cite{zhang2021weakly} designed a regional feature alignment module to address the spatial weak misalignment issue between modalities, adaptively aligning cross-modal regional features by predicting the positional offsets between modalities. He et al.~\cite{he2023multispectral} designed a cross-modal conflict correction module to mitigate semantic conflicts arising from the heterogeneity between modalities by aggregating adjacent similar patches within a modality and utilizing contextual information between modalities to alleviate their heterogeneity. 
Unlike these works, in this paper, we propose leveraging the powerful semantic representation capabilities of large language models to guide the model in progressively aligning semantic and spatial information, achieving significantly better performance than our baseline.

\subsection{Cross-Modal Image Alignment.} 

In multimodal tasks, images from different modalities often exhibit differences in scale, angle, and spatial position. After pre-alignment, simple scale and angle differences are well addressed. However, there are too many factors causing spatial misalignment, making it difficult to achieve complete alignment through simple geometric transformations alone. To address this issue, some researchers~\cite{zhang2021multimodal} have adopted pixel-level pre-alignment methods, which explicitly adjust pixel coordinates using a global mapping function learned from multimodal images to achieve strict pre-alignment. However, this approach often requires pre-training and additional supervision. To adapt to an end-to-end approach, Zhang et al.~\cite{zhang2021weakly} inserted a region feature alignment module into the network to predict the positional offsets of the regions to be fused and adjusted the region features for alignment. Yuan et al.~\cite{yuan2022translation} decomposed the misalignment issue into deviations in position, size, and angle. And explicitly performed feature alignment by applying translation, rotation, and scaling operations on the feature maps of candidate regions using the predicted inter-object deviations. C$^{2}$Former~\cite{yuan2024c2former} utilized the ability of Transformers to model pairwise associations between different features, designing a cross-modal cross-attention module to learn the cross-attention relationships between modalities and calibrate complementary features. Chen et al.~\cite{chen2024weakly} constructed a cross-modal common subspace to decouple modality-invariant features and utilized spatial consistency to estimate cross-modal offsets, guiding adaptive feature alignment.
However, existing multimodal drone object detection methods often overlook the impact of semantic gaps between modalities, making it difficult for them to achieve accurate spatial alignment, thereby limiting detection performance.

\subsection{Large Language Model Guided Vision Tasks}

In recent years, with the rapid development of large language models, significant progress has been made in tasks such as text generation, understanding, and reasoning. Large Language Models such as the GPT series~\cite{radford2018GPT1, radford2019GPT2, brown2020GPT3, openai2023gpt4}, BERT~\cite{kenton2019bert}, LLaMA~\cite{touvron2023llama} and T5~\cite{raffel2020exploring} have been widely used in the Natural Language Processing (NLP) field. Due to their strong text understanding and generation capabilities, large language models provide rich semantic information and contextual guidance for visual tasks. As a result, some researchers begin to explore the application of large language models in vision tasks. 
Li et al.~\cite{li2023clip} fully leverage Multimodal Large Language Models cross-modal descriptive capability by assigning a set of learnable prompts for each ID, inputting these prompts into the text encoder to form ambiguous descriptions of the images, thereby facilitating better visual representations. Zhang et al.~\cite{zhang2024groundhog} integrated LLM with holistic segmentation, enhancing reasoning and diagnostic capabilities in visual tasks. Wang et al.~\cite{wang2024visionllm} argue that visual foundation models struggle to handle open-ended visual tasks and therefore propose a framework that leverages large language models to perform open-ended visual tasks. Wu et al.~\cite{wu2024vfm} introduce large language models into the vehicle object detection task to encode attribute information and learn unified attribute representations, thereby mitigating the semantic gap between multimodal features. Inspired by these works, we introduce large language models into multimodal object UAV detection, leveraging their powerful semantic representation capabilities to guide the model in progressively aligning the semantic and spatial information of multimodal features.

%% file: sec_arxiv/3_meth.tex
\section{Methodology}
\label{sec:meth}

\subsection{Overview}

\begin{figure*}
\centering
\includegraphics[width=\textwidth]{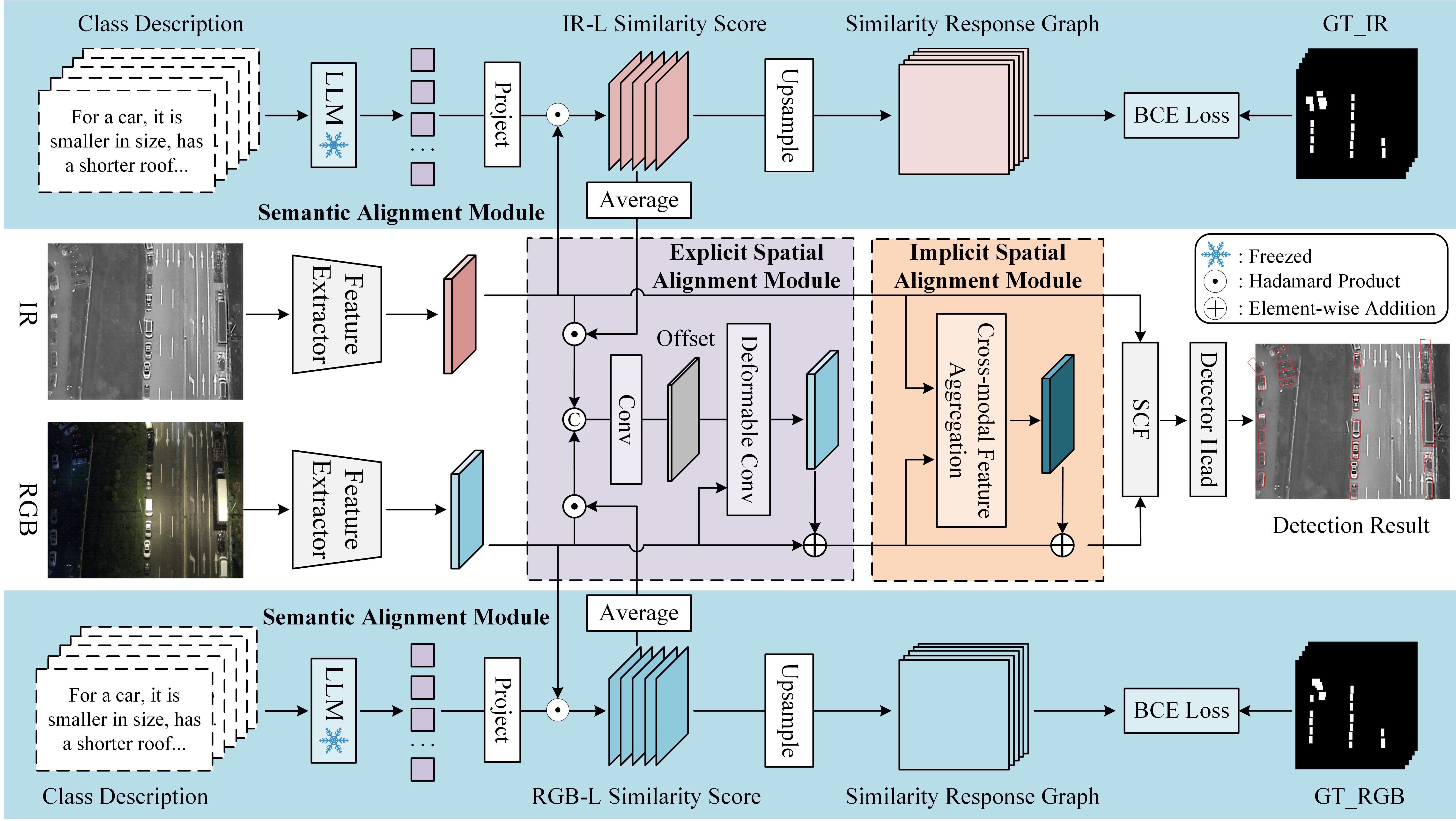}
\caption{The framework diagram of our proposed Large Language Model guided Progressive feature Alignment Network (LPANet) for multi-modal UAV object detection} 
\label{framework}
\end{figure*}

As shown in Figure~\ref{framework}, we propose the Large Language Model guided Progressive feature Alignment Network framework (LPANet), which aims to leverage the semantic features of each object category extracted by a large language model to guide semantic and spatial alignment of multimodal features. 
This framework inputs RGB-IR image pairs into a dual-stream backbone network to extract multi-scale visual features. 
Meanwhile, the pre-trained large language model MPNet~\cite{song2020mpnet} is introduced to encode the fine-grained textual descriptions generated with the assistance of ChatGPT, thereby progressively guiding the semantic and spatial alignment of multimodal features using the obtained semantic features.
Specifically, we first design the Semantic Alignment Module (SAM), which maximizes the similarity between the semantic features of objects and multimodal visual features, bringing them closer in a shared space to achieve semantic alignment of multimodal features.
Secondly, we design the Explicit Spatial alignment Module (ESM), which enhances the object regions in multimodal visual features using the semantic relationships obtained from SAM, thereby estimating a more accurate spatial offset that guides deformable convolution to achieve coarse spatial alignment between RGB-IR modalities.
Finally, the Implicit Spatial alignment Module (ISM) aggregates key features from adjacent regions using cross-modal correlation, achieving implicit spatial alignment between RGB-IR modalities. 

\subsection{Semantic Alignment Module}

Existing multimodal~\cite{sun2022drone, yuan2022translation, yuan2024c2former} UAV object detection algorithms often overlook the semantic gap between modalities, making the accurate alignment of both semantic and spatial features difficult. To address this issue, we design the SAM, which introduces large language model encoded textual descriptions of object categories to extract semantic features. Then, we bring multimodal visual features and semantic features of each object closer in a shared space, thereby alleviating the semantic gap between modalities and providing the model with richer semantic information.

Specifically, give the RGB-IR image pair $I_{RGB} \in \mathbb{R}^{H \times W \times 3}$ and $I_{IR} \in \mathbb{R}^{H \times W \times 1}$, we first pass them through their corresponding pyramid feature extractors to obtain the RGB features $F_{RGB}$ and the IR features $F_{IR}$. These feature extractors have the same network architecture, but their parameters are different.
Meanwhile, we leverage ChatGPT to assist in generating distinctive text descriptions for each category. Specifically, we first prompt ChatGPT with category and scene information to obtain features that are conducive to distinguishing between categories. After manually selecting prominent features, we utilize the large language model to generate a set of object category description texts $T={t_1, t_2, ..., t_n}$. As shown in Table~\ref{ablation_text}, we provide the description texts corresponding to each object category. Subsequently, we employ the pre-trained large language model MPNet~\cite{song2020mpnet} as a text encoder to encode the category description texts $t_i$, obtaining semantic embeddings $p^{i}_{emb} \in \mathbb{R}^{1 \times 768}, i \in \{1, 2, ..., n\}$. By concatenating all the semantic embeddings, we obtain the semantic feature $F_t$.

We feed the semantic, RGB, and IR features into the semantic alignment module together and use three independent 256-dimensional linear projection layers ($P_{\theta_{RGB}}()$, $P_{\theta_{IR}}()$, $P_{\theta_{t}}()$ respectively) to project these features into the common dimension. We then perform matrix multiplication between the projected RGB features $\overline{F}_{RGB}$ and IR features $\overline{F}_{IR}$, and the semantic features $\overline{F}_t$ to obtain the RGB-Semantic and IR-Semantic similarity score, denoted as $S_{RGB}$ and $S_{IR}$, respectively.

To bring each object's semantic features and multimodal visual features closer in a shared space, we generate the RGB mask map $m^{i}_{RGB}, i \in \{1, 2, ..., n\}$ and IR mask map $m^{i}_{IR}, i \in \{1, 2, ..., n\}$ for each category using the ground truth labels. For the similarity score $s_{RGB}$ and $s_{IR}$, we use bilinear interpolation to upsample them to the original image size, obtaining the similarity response maps $S_{RGB}$ and $S_{IR}$. Finally, we compute the binary cross-entropy loss between each category similarity maps and their corresponding ground truth mask maps, which can be expressed as:
\begin{equation}
\label{loss1}
L_{sa} = BCE( m_{RGB},S_{RGB}) + BCE( m_{IR},S_{IR}),
\end{equation} 
where $BCE()$ represents the binary cross-entropy loss.

\begin{table}[!htp]
\centering 

\resizebox{3.3in}{!}{
\begin{tabular} {c|p{2.5in}}
\hline 
\textbf{Class Name} & \textbf{Class Description}  \\
\hline 
 Car& For a car, the size is smaller, the roof is shorter in length, and there is no cargo compartment. \\

 \hline
 
 Van& For a van, the size is relatively small but larger than a car, smaller than a truck, and has a cargo area integrated with the cab. \\

\hline
 
 Truck& For a truck, the cab is smaller, and the cargo area is long and flat, but slightly shorter than that of a freight car. \\

\hline
 
 Freight car& For a freight car, it has the largest volume and the longest length, with a larger cargo area compared to a truck. \\

\hline
 
 Bus& For a bus, the body is continuous and long, without a separation between the cab and the rest of the vehicle. \\
 
\hline
\end{tabular} }
\caption{Display of category names and their corresponding text descriptions.} 
\label{ablation_text} 
\end{table}
 
\subsection{Explicit Spatial Alignment Module}

We obtain the semantically aligned RGB and IR features from the SAM. However, inherent differences between modalities lead to spatial misalignment between the RGB-IR image pairs. Since detection models are sensitive to positional information, spatial misalignment can negatively impact detection performance. Therefore, to better align spatial positions between modalities, we design the ESM.

In the ESM, we first calculate the spatial differences between the semantically aligned features $\overline{F}_{RGB}$ and $\overline{F}_{IR}$ to simulate the offset and then use deformable convolution for coarse spatial alignment. Common methods for calculating feature differences include subtraction and concatenation. The subtraction method calculates the difference by directly subtracting the features, while the concatenation method concatenates the two features along the channel dimension and estimates the difference through the model.

To mitigate the interference of background noise in spatial offset estimation, we introduce the semantic relationships obtained in SAM to enhance the features of the object region while suppressing background noise. Specifically, we perform element-wise multiplication the semantically aligned features $\overline{F}_{RGB}$ and $\overline{F}_{IR}$ with the similarity scores $S_{RGB}$ and $S_{RI}$ calculated in SAM, obtaining the enhanced object features $\tilde{F}_{RGB}$ and $\tilde{F}_{IR}$. Then, we use the concatenation method to concatenate the features $\tilde{F}_{RGB}$ and $\tilde{F}_{IR}$ along the channel dimension. The concatenated features $F_{all}$ are fed into a convolutional layer with a 3x3 kernel to estimate the feature-level offset $\phi$ of RGB relative to IR.
\begin{equation}
\label{srgb}
\phi = Conv(ConCat((\overline{F}_{RGB} \cdot S_{RGB}),(\overline{F}_{IR} \cdot S_{IR}))),
\end{equation} 
we feed the estimated spatial offsets $\phi$, and RGB features $\overline{F}_{RGB}$ into a deformable convolution, which, based on conventional convolution operations, dynamically adjusts the sampling positions of the convolutional kernel using the spatial offsets, resulting in spatially deformed RGB features $F^{a}_{RGB}$ to achieve coarse alignment between modalities.

\begin{figure}
\centering
\includegraphics[width=3.3in]{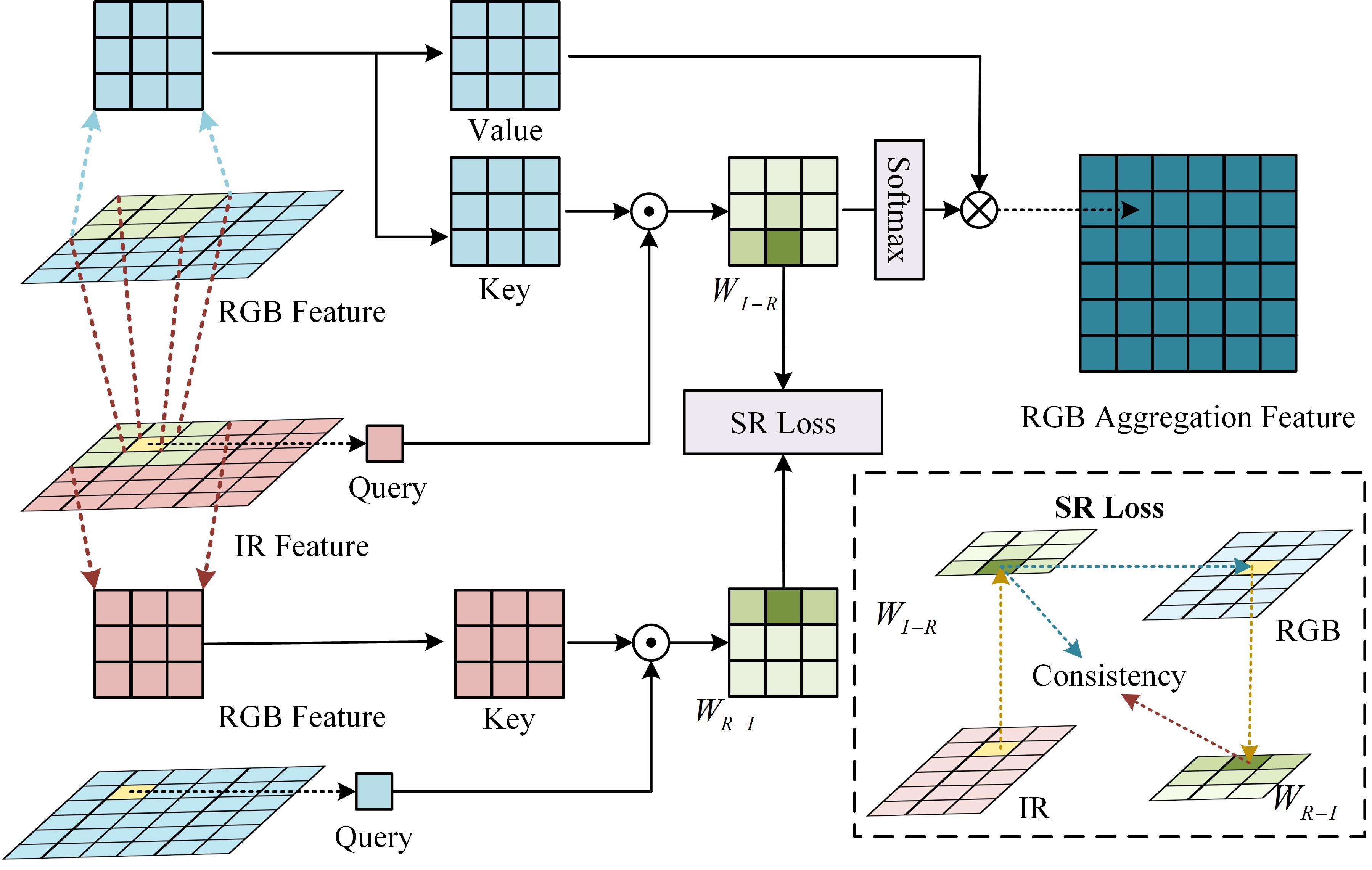}
\caption{The structures of Implicit Spatial alignment Module. The bottom-right corner shows an example of how we calculate the Symmetric Consistency Loss (SR Loss).} 
\label{SIM}
\end{figure}

\subsection{Implicit  Spatial Alignment Module}

Since the estimation of offsets is influenced by numerous factors, ESM can only achieve coarse spatial alignment. To further refine the spatial features, we propose an Implicit  Spatial Alignment Module (ISM), which aggregates key information from neighboring regions, thereby implicitly performing cross-modal spatial alignment. 
As shown in Figure~\ref{SIM}, the IR features $\overline{F}_{IR}$ and the RGB features $F^{a}_{RGB}$ output by ESM are fed into this module. We use each patch of the IR features $\overline{F}_{IR}$ as the query and sample the RGB features $F^{a}_{RGB}$ using a 3x3 sliding window. The sampled region features are concatenated and used as the key and value for input into the cross-modal attention mechanism. We compute the dot product between $\overline{F}_{IR}$ and $F^{w}_{RGB}$ to obtain the similarity matrix $w_{I-R}$, and then apply softmax to obtain the attention weight $\overline{w}_{I-R}$. This process can be represented as:
\begin{equation}
\label{w1}
\overline{w}_{I-R} = Softmax(\frac{\overline{F}_{IR} \cdot{F^{w}_{RGB}}^T}{\sqrt{d}}),
\end{equation} 
where $d$ represents the scaling factor. To aggregate the key features within the window region, we use attention weights $\overline{w}_{I-R}$  to perform a weighted aggregation of the RGB features $F^{w}_{RGB}$, obtaining the implicitly aligned RGB features $\tilde{F}^{a}_{RGB}$. 
To constrain the implicit alignment process, we swap the RGB and IR features and repeat the above steps to compute the cross-modal attention, resulting in the similarity matrix $w_{R-I}$. Then, we apply a Symmetric Consistency Loss (SR Loss) to enforce consistency between the similarity matrix $w_{I-R}$ and $w_{R-I}$. 

Specifically, we first extract the similarity matrix $w^{i}_{I-R}$ of the $i$-th patch $p^{IR}_i$ from the overall similarity matrix $w_{I-R}$, and then extract the highest similarity value $v^{i}_{I-R}$ from $w^{i}_{I-R}$ along with the corresponding index $j$ of the RGB feature patch. Next, we locate the similarity matrix $w^{i}_{R-I}$ for the $j$-th patch $p^{RGB}_j$ from the similarity matrix $w_{R-I}$, and extract its similarity value $v^{i}_{R-I}$ with the IR feature patch $p^{IR}_i$. By repeating this process for each image patch in the IR features, we obtain two corresponding similarity matrices $V_{I-R}$ and $V_{R-I}$. Therefore, the symmetric consistency loss can be simply represented as:
\begin{equation}
\label{losskl}
L_{sc} = KL(V_{I-R},V_{R-I}),
\end{equation} 
where $KL$ represents the Kullback-Leibler Divergence. 
Finally, we feed the aligned RGB features $\tilde{F}^{a}_{RGB}$ and IR features $\overline{F}_{IR}$ into the same fusion module as the baseline. By incorporating cross-modal consistency information, we capture rich semantic features for fusion. The fused features are then passed into the detector head to predict the final detection results.

\begin{table*}[!htp]
\centering 

\vspace{5pt} 
\resizebox{\textwidth}{!}{
\begin{tabular}{c|l|cccccc|cccccc}  
\hline
\multirow{2}{*}{\textbf{Modality}}
&\multirow{2}{*}{\textbf{Detectors}}
&\multicolumn{6}{c|}{\textbf{DroneVehicle Val Set}} 
&\multicolumn{6}{c}{\textbf{DroneVehicle Test Set}} \\
& &\textbf{Car}  &\textbf{Truck} &\textbf{Freight car} &\textbf{Bus}  &\textbf{Van} & \textbf{$mAP_{0.5}$} &\textbf{Car}  &\textbf{Truck} &\textbf{Freight car} &\textbf{Bus}  &\textbf{Van} & \textbf{$mAP_{0.5}$} \\
\hline 
\multirow{6}{*}{\raggedright \textbf{RGB}}
&RetinaNet~\cite{ross2017focal} &78.5 &34.4 &24.1 &69.8 &28.8 &47.1    &67.5 &28.2 &13.7 &62.1 &19.3 &38.1 \\
&Faster R-CNN~\cite{ren2016faster} &79.0 &49.0 &37.2 &77.0 &37.0 &55.9    &67.9 &38.6 &26.3 &67.0 &23.2 &44.6  \\
&Oriented R-CNN~\cite{xie2021oriented} &80.1 &53.8 &41.6 &85.4 &43.3 &60.8    &- &- &- &- &- &-  \\
&S$^{2}$ANet~\cite{han2021align} &80.0 &54.2 &42.2 &84.9 &43.8 &61.0    &- &- &- &- &- &-  \\
&RoITransformer~\cite{ding2019learning} &61.6 &55.1 &42.3 &85.5 &44.8 &61.6    &68.1 &44.2 &29.1 &70.6 &27.6 &47.9  \\
&YOLOv5~\cite{jia2021llvip} &78.6 &55.3 &43.8 &87.1 &46.0 &62.1    &76.2 &48.9 &35.5 &68.9 &40.4 &54.0  \\
\hline 
\multirow{6}{*}{\raggedright \textbf{IR}}
&RetinaNet~\cite{ross2017focal} &88.8 &35.4 &39.5 &76.5 &32.1 &54.5    &79.9 &32.8 &28.1 &67.3 &16.4 &44.9  \\
&Faster R-CNN~\cite{ren2016faster} &89.4 &53.5 &48.3 &87.0 &42.6 &64.2    &88.6 &42.5 &35.2 &77.9 &28.5 &54.6  \\
&Oriented R-CNN~\cite{xie2021oriented} &89.8 &57.4 &53.1 &89.3 &45.4 &67.0    &- &- &- &- &- &-  \\
&S$^{2}$ANet~\cite{han2021align} &89.9 &54.5 &55.8 &88.9 &48.4 &67.5     &- &- &- &- &- &- \\
&RoITransformer~\cite{ding2019learning} &90.1 &60.4 &58.9 &89.7 &52.2 &70.3    &88.9 &51.5 &41.5 &79.5 &34.4 &59.2  \\
&YOLOv5~\cite{jia2021llvip} &90.0 &59.5 &60.8 &89.5 &53.8 &70.7    &89.3 &59.9 &49.5 &86.7 &42.1 &65.5  \\
&DTNet~\cite{zhang2024dtnet} &90.2 &78.1 &67.9 &89.2 &65.7 &78.2    &- &- &- &- &- &- \\
\hline 
\multirow{11}{*}{\raggedright \textbf{RGB+IR}}
&UA-CMDet~\cite{sun2022drone} &- &- &- &- &- &-    &87.5 &60.7 &46.8 &87.1 &38.0 &64.0 \\
&Halfway Fusion~\cite{liu2016multispectral} &90.1 &62.3 &58.5 &89.1 &49.8 &70.0    &- &- &- &- &- &- \\
&CIAN~\cite{zhang2019cross} &90.1 &63.8 &60.7 &89.1 &50.3 &70.8    &- &- &- &- &- &- \\
&AR-CNN~\cite{zhang2021weakly} &90.1 &64.8 &62.1 &89.4 &51.5 &71.6    &- &- &- &- &- &- \\
&MBNet~\cite{zhou2020improving} &90.1 &64.4 &62.4 &88.8 &53.6 &71.9    &- &- &- &- &- &- \\
&TSFADet~\cite{yuan2022translation} &89.9 &67.9 &63.7 &89.8 &54.0 &73.1    &89.2 &72.0 &54.2 &88.1 &48.8 &70.4 \\
&C$^{2}$Former~\cite{yuan2024c2former}  &- &- &- &- &- &-   &90.2 &68.3 &64.4 &89.8 &58.5 &74.2     \\
&SLBAF-Net~\cite{cheng2023slbaf} &90.2 &72.0 &68.6 &89.9 &59.9 &76.1    &- &- &- &- &- &- \\
&CALNet~\cite{he2023multispectral} &90.3 &73.7 &68.7 &89.7 &59.7 &76.4    &90.3 &76.2 &63.0 &89.1 &58.5 &75.4 \\
&OAFA~\cite{chen2024weakly} &90.3 &76.8 &73.3 &\textbf{90.3} &66.0 &79.4    &- &- &- &- &- &- \\
&Ours &\textbf{90.4} &\textbf{80.6} &\textbf{76.3} &90.1 &\textbf{67.5} &\textbf{81.0}    &\textbf{90.4} &\textbf{78.0} &\textbf{65.0} &\textbf{89.5} &\textbf{65.4} &\textbf{77.7} \\
\hline
\end{tabular}   }
\caption{Experimental results of and other detection algorithms on DroneVehicle object detection datasets.} 
\label{detection} 
\end{table*}

\begin{table*}[!htp]
\centering  
\vspace{5pt} 
\resizebox{\textwidth}{!}{
\begin{tabular}{c|l|ccccccccc|c}  
\hline
\textbf{Modality} &\textbf{Detectors} &\textbf{Car}  &\textbf{Truck} &\textbf{Tractor} &\textbf{Camping Car}  &\textbf{Van} &\textbf{Pick-up}  &\textbf{Boat} &\textbf{Plane} &\textbf{Others}  & \textbf{$mAP_{0.5}$} \\
\hline 
\multirow{5}{*}{\raggedright \textbf{RGB}}
&RetinaNet~\cite{ross2017focal} &48.9 &16.8 &15.9 &21.4 &5.9 &37.5 &4.4 &21.2 &14.1   &20.7    \\
&S$^{2}$ANet~\cite{han2021align} &74.5 &47.3 &55.6 &61.7 &32.5 &65.1 &16.7 &7.1 &39.8   &44.5 \\
&Faster R-CNN~\cite{ren2016faster} &71.4 &54.2 &61.0 &70.5 &59.5 &67.6 &52.3 &77.1 &40.1 &61.5 \\
&RoITransformer~\cite{ding2019learning} &77.3 &56.1 &64.7 &73.6 &60.2 &71.5 &56.7 &85.7 &42.8   &65.4 \\
&Oriented R-CNN~\cite{xie2021oriented} &77.6 &59.7 &62.8 &76.7 &60.9 &72.3 &60.1 &84.0 &43.6 &66.4 \\
\hline 
\multirow{5}{*}{\raggedright \textbf{IR}}
&RetinaNet~\cite{ross2017focal} &44.2 &15.3 &9.4 &17.1 &7.2 &32.1 &4.0 &33.4 &5.7 &18.7    \\
&S$^{2}$ANet~\cite{han2021align} &73.0 &39.2 &41.9 &59.2 &32.3 &65.6 &13.9 &12.0 &23.1 &40.0 \\
&Faster R-CNN~\cite{ren2016faster} &71.6 &49.1 &49.2 &68.1 &57.0 &66.5 &35.6 &71.6 &29.5   &55.4 \\
&RoITransformer~\cite{ding2019learning} &76.1 &51.7 &51.9 &71.2 &64.3 &70.7 &46.9 &83.3 &28.3   &60.5 \\
&Oriented R-CNN~\cite{xie2021oriented} &77.0 &55.0 &47.5 &73.6 &63.2 &72.2 &49.4 &79.6 &30.5   &60.9 \\
\hline 
\multirow{4}{*}{\raggedright \textbf{RGB+IR}}
&C$^{2}$Former+S$^{2}$ANet~\cite{yuan2024c2former} &76.7 &52.0 &59.8 &63.2 &48.0 &68.7 &43.3 &47.0 &41.9   &55.6 \\
&CMAFF+Oriented R-CNN~\cite{qingyun2022cross} &81.7 &58.8 &68.7 &78.4 &68.5 &76.3 &66.0 &72.7 &51.5   &69.2 \\
&CALNet~\cite{he2023multispectral} &82.3 &66.7 &73.0 &75.4 &65.5 &78.2 &56.6 &90.5 &47.0   &70.6 \\

&Ours &\textbf{82.4} &\textbf{70.6} &\textbf{77.0} &\textbf{78.4} &\textbf{75.4} &\textbf{80.0} &\textbf{69.0} &\textbf{97.2} &\textbf{56.7} &\textbf{76.4} \\
\hline
\end{tabular}   }
\caption{Experimental results of and other detection algorithms on VEDAI datasets.} 
\label{vedai}
\end{table*}

\subsection{Model Training}

During model training, in addition to the loss functions introduced above, we retain the existing loss functions from the baseline model, including the bounding box loss $L_{box}$, objectness loss $L_{obj}$, classification loss $L_{cls}$, and angle loss $L_{theta}$. The total loss function can be expressed as:
\begin{equation}
\label{lossall}
L = L_{box} + L_{obj} + L_{cls} + L_{theta} + L_{sa} + L_{sc}.
\end{equation} 
Semantic alignment is crucial for ESM to predict spatial offsets accurately, so we divide training into two stages. In the first stage, only the SAM and ISM are trained, providing semantically aligned RGB and IR features, which help ESM predict spatial offsets ($\phi$) more precisely. In the second stage, all modules are trained, with the first-stage weights used for initialization. This prior training significantly reduces the semantic gap between RGB and IR features, enabling more accurate spatial offset prediction and cross-modal spatial alignment.

%% file: sec_arxiv/4_exper.tex
\section{Experiments}
\label{sec:exper}

\subsection{Datasets and Evaluation Metrics} 

\textbf{Datasets. }  
In this work, we validate our method on two multimodal object detection datasets, DroneVehicle~\cite{li2023clip} and VEDAI~\cite{razakarivony2016vehicle}. 

\textbf{DroneVehicle}~\cite{li2023clip} is a large-scale multimodal UAV detection dataset. The dataset consists of 28439 pairs of pre-registered RGB-IR image pairs. It covers various perspectives, altitudes, scenes, and lighting conditions. The training, test, and validation set contain 17990, 8980, and 1469 image pairs, respectively. The dataset provides oriented bounding box annotations for five categories: car, truck, bus, van, and freight car. 

\textbf{VEDAI}~\cite{razakarivony2016vehicle} is a multimodal small object detection dataset that covers diverse urban and rural scenes. It provides oriented bounding box annotations for nine categories, including Car, Van, and Boat. The dataset consists of 1246 pairs of RGB-IR image pairs, available in two resolutions: $512 \times 512$ and $1024 \times 1024$. In this work, we use the $512 \times 512$ resolution and follow the ten-fold cross-validation protocol proposed in the original paper~\cite{razakarivony2016vehicle} to evaluate our model.

\begin{figure}
\centering
\includegraphics[width=3.3in]{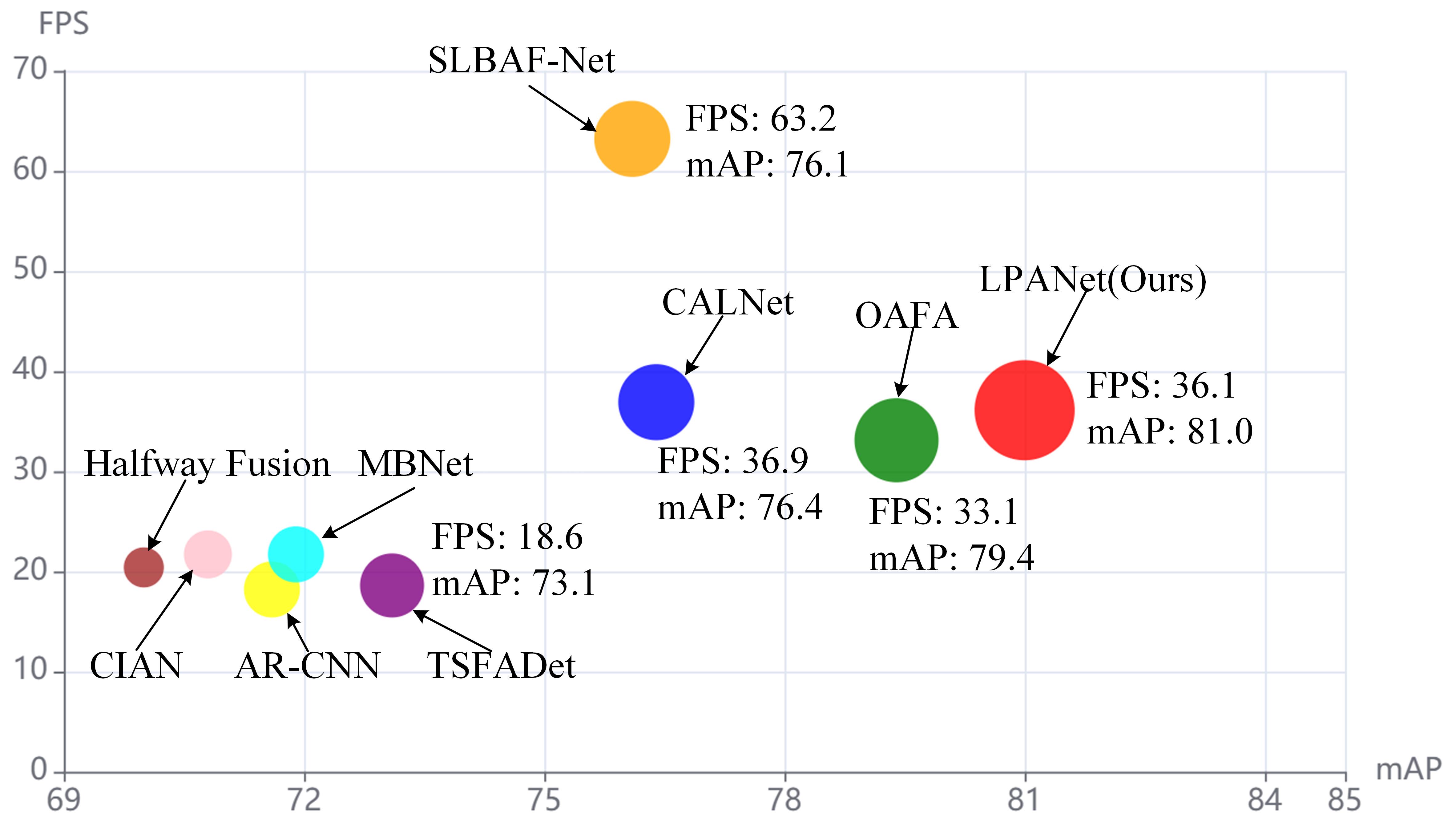}
\caption{Comparison of speed and accuracy on the DroneVehicle. Speed is compared using FPS.} 
\label{fps}
\end{figure}

\textbf{Evaluation Metrics. }  
In this work, we use the mean Average Precision (mAP) metric at an Intersection over Union (IoU) threshold of 0.5 to evaluate the accuracy of multimodal object detection.

\subsection{Implementation Details}  

Our model training is carried out in two stages. In the first stage, we initialize the model by loading the pre-trained weights of the backbone network, setting the learning rate to 0.035. In the second stage, we load the parameters trained in the first stage and reduce the learning rate to 0.02. The batch size is set to 4, momentum is set to 0.843, and weight decay is set to 0.00036 during both training stages. We employ SGD as the optimizer for model training, using 50 epochs per stage, totaling 100 epochs. All experiments are implemented in Python using the deep learning framework PyTorch. The model training and testing are conducted on an NVIDIA RTX 4090.

\subsection{Comparison Results} 

\textbf{Comparisons on DroneVehicle Dataset. }  
We compare our method with other approaches to the validation and test sets of the DroneVehicle dataset. As shown in Table~\ref{detection}, we compare our method with 17 state-of-the-art methods, including six unimodal and eleven multimodal methods.
Among the six unimodal methods, we assess the detection performance of models trained solely on RGB or IR images. The results indicate that training on the IR modality significantly outperforms training on the RGB modality, underscoring the advantage of IR images in all-weather object detection tasks. 
For the DroneVehicle validation set, our baseline method CALNet achieves an mAP of $76.4\%$, with category-wise precision values of $90.3\%$, $73.7\%$, $68.7\%$, $89.7\%$, and $59.7\%$, respectively. 
Compared to baseline method, our method improves the mAP by $4.6\%$, and also increases precision each object category by $0.1\%$, $6.9\%$, $7.6\%$, $0.4\%$, and $7.8\%$, respectively. Moreover, our method surpasses the second-best method by $1.6\%$ in mAP. Notably, our method demonstrates significant improvement in distinguishing the easily confused categories of Truck, Van, and Freight Car compared to the baseline. This improvement is likely attributed to the introduction of fine-grained text descriptions for each category, which provide the model with rich semantic priors. On the DroneVehicle test set, our method improves the mAP by $2.3\%$ compared to the baseline, achieving the best performance. These experimental results and comparisons clearly demonstrate the superiority of our model.

\textbf{Comparisons on VEDAI Dataset.}
To assess the generalization ability of our method in scenarios with a more diverse set of categories, we compare it with other detection algorithms on the VEDAI dataset, as shown in Table~\ref{vedai}. Our baseline method, CALNet, achieves an mAP of $70.6\%$, with per-class accuracies of $82.3\%$, $66.7\%$, $73.0\%$, $75.4\%$, $65.5\%$, $78.2\%$, $56.6\%$, $90.5\%$, and $47.0\%$, respectively. Notably, our method outperforms the mAP by $5.8\%$ compared to the baseline model, while the per-class accuracies increase by $0.1\%$, $3.9\%$, $4.0\%$, $3.0\%$, $9.9\%$, $1.8\%$, $12.4\%$, $6.7\%$, and $9.7\%$, achieving the best performance. These experimental results and comparative analyses clearly demonstrate the generalization ability of our model.

\textbf{Comparsion on Speed.}
As shown in Figure~\ref{fps}, we compare the inference speed of our method with other detectors. During the speed test, we keep the settings the same as those of the other methods, setting the batch size to 1 and the input size to $640 \times 640$. From the results, our method achieves 36.1 FPS, which outperforms most existing UAV multimodal object detection algorithms. Meanwhile, compared to the baseline, our method only experiences a decrease of 0.8 FPS, while improving accuracy by $4.6\%$.
\begin{figure}
\centering
\includegraphics[width=3.3in]{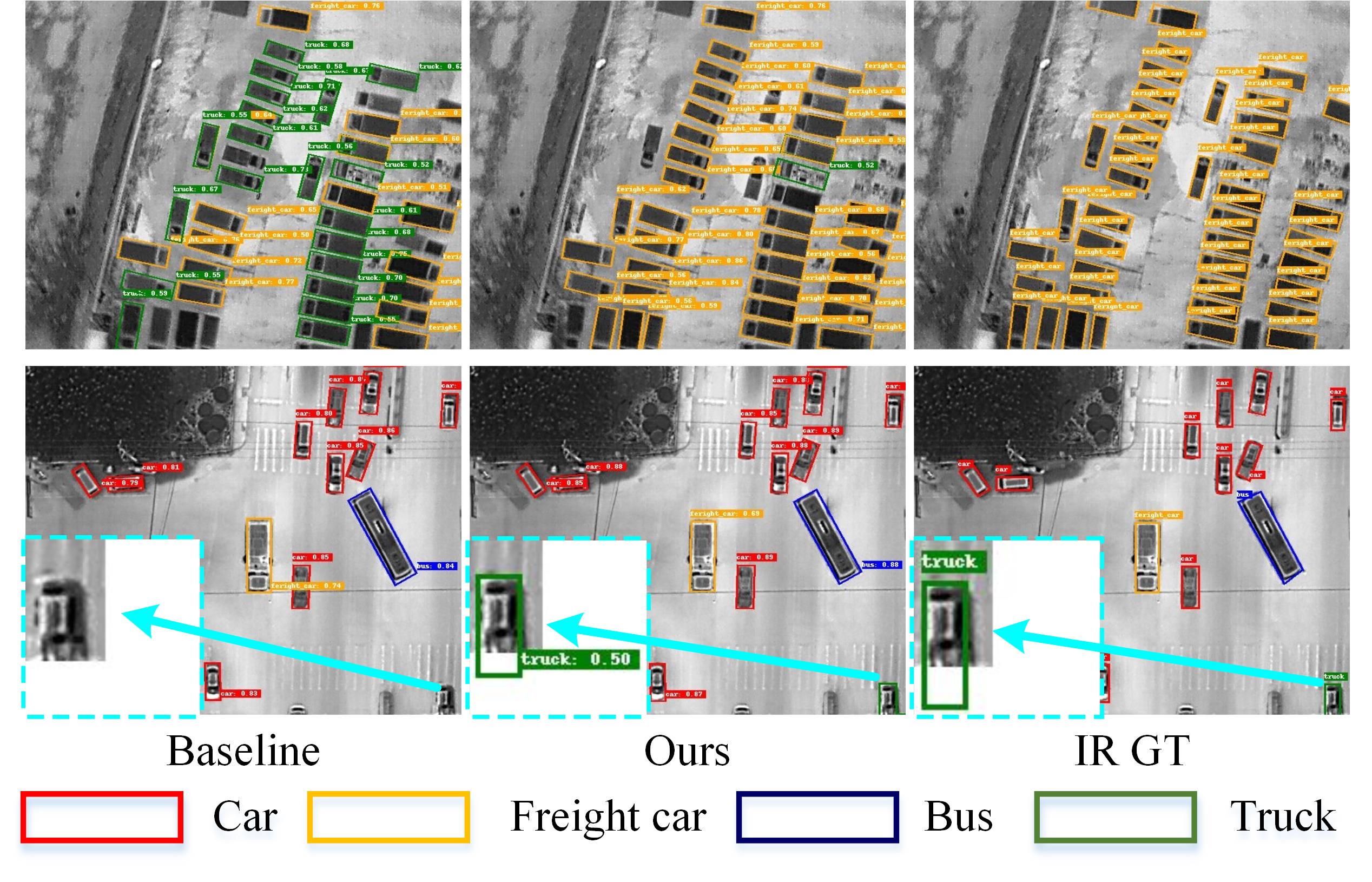}
\caption{Visualization of detection results on the DroneVehicle dataset, with different color boxes representing different categories.} 
\label{detection_result}
\end{figure} 
\begin{figure*}
\centering
\includegraphics[width=\textwidth]{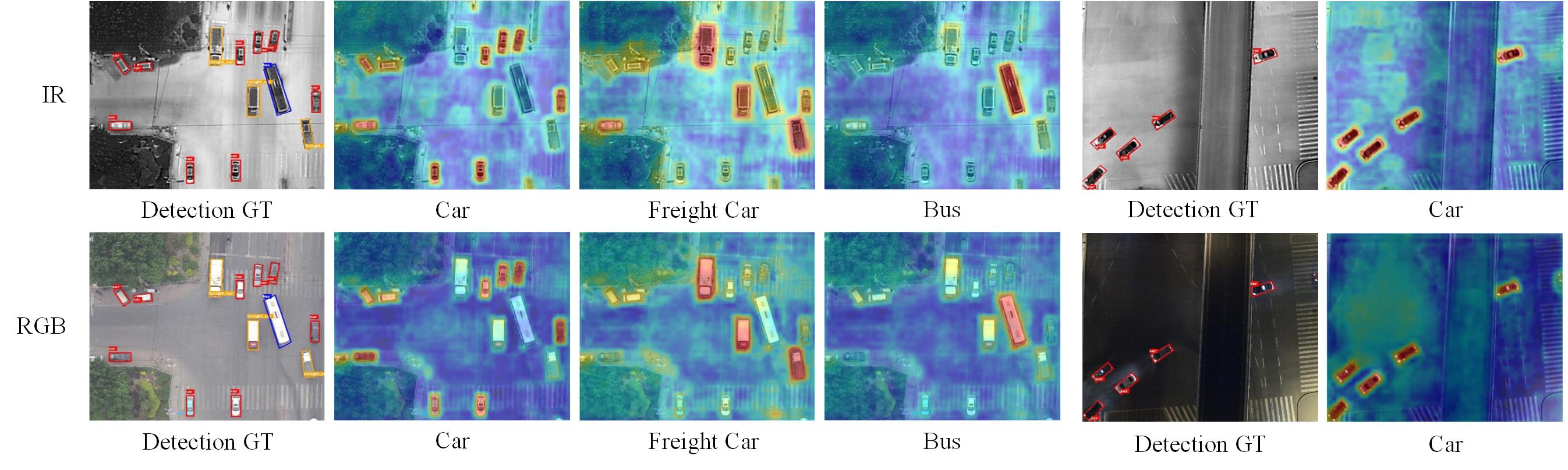}
\caption{Visualization of similarity response maps between text descriptions of various categories and multimodal visual features.} 
\label{VT}
\end{figure*} 
\begin{figure}
\centering
\includegraphics[width=3.3in]{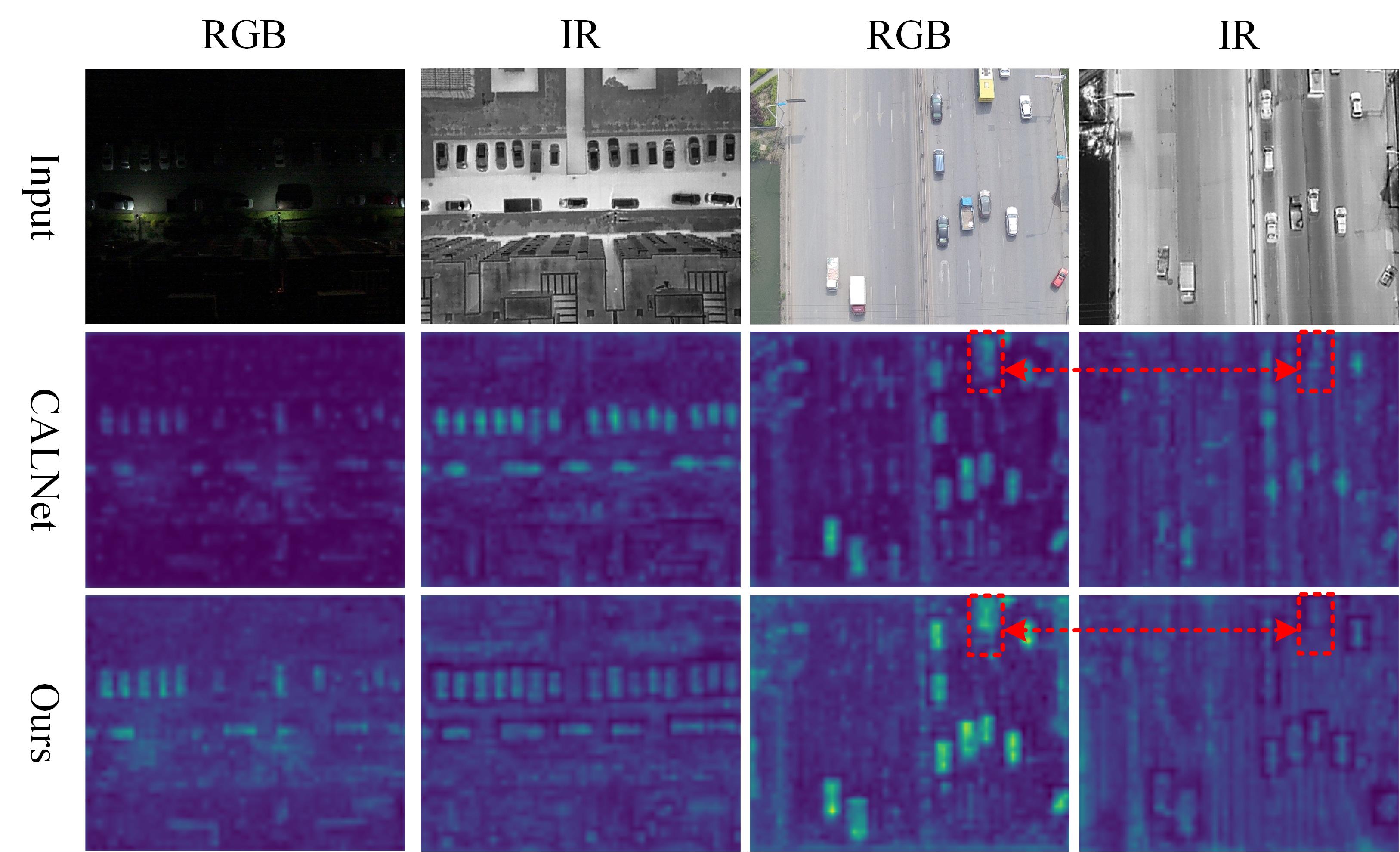}
\caption{Visualization results of feature maps for RGB and IR.} 
\label{featuremap}
\end{figure} 

\subsection{Ablation Study} 

\textbf{Effects of Semantic Alignment Module.} 
In this paper, we propose the SAM, which aligns semantic information across modalities by maximizing the similarity between each object's semantic features and multimodal visual features in a shared space. As shown in Table~\ref{ablation_model}, incorporating the SAM into the baseline model results in a $1.8\%$ improvement in mAP, with almost no impact on computational efficiency. Moreover, the accuracy of the three easily confused categories, Truck, Freight car, and Van, increased by $2.2\%$, $3.3\%$, and $1.6\%$, respectively. These experimental results and comparisons comprehensively validate the effectiveness of the proposed semantic alignment module.
\begin{table}[!htp]
\centering  
\resizebox{3.3in}{!}{
\begin{tabular}{c|ccccccc}  
\hline 
\textbf{Method} &\textbf{Car}  &\textbf{Truck} &\textbf{F.car} &\textbf{Bus}  &\textbf{Van} & \textbf{$mAP_{0.5}$} &\textbf{FLOPs} \\
\hline 
Baseline  &90.3 &73.7 &68.7 &89.7 &59.7 &76.4 &171.53G \\
+SAM &90.2 &75.9 &72.0 &89.8 &61.3 &77.8 &171.58G \\
+ISM  &90.4 &78.0 &74.6 &90.0 &63.2 &79.3 &184.10G \\
+ESM  &\textbf{90.4} &\textbf{80.6} &\textbf{76.3} &\textbf{90.1} &\textbf{67.5} &\textbf{81.0} &184.63G \\
\hline
\end{tabular} }
\caption{Ablation study on DroneVehicle dataset. F.car denotes freight car.} 
\label{ablation_model}
\end{table}

\textbf{Effects of Implicit Spatial alignment Module.}
To address the spatial misalignment between modalities, we propose the ISM. This module leverages cross-modal correlation to aggregate key features from adjacent regions, thereby achieving implicit alignment across modalities. To further enhance the aggregate key features, we introduce a symmetric consistency loss as a constraint. As shown in Table~\ref{ablation_model}, after adding the ISM, the mAP increases to $79.3\%$, representing a $1.5\%$ improvement over using only the SAM. These experimental results and comparisons fully validate the effectiveness of the proposed ISM in addressing spatial misalignment issues. Due to the large amount of attention computation involved in the ISM, FLOPs increase by 12.52G. However, this module effectively alleviates the weak alignment issue between modalities and improves detection performance. Therefore, we consider the additional computational cost to be worthwhile.

\begin{table}[!htp]
\centering 
\resizebox{3.3in}{!}{
\begin{tabular}{c|cccccc}  
\hline 
\textbf{Method} &\textbf{Car}  &\textbf{Truck} &\textbf{Freight car} &\textbf{Bus}  &\textbf{Van} & \textbf{$mAP_{0.5}$} \\
\hline 
BERT~\cite{kenton2019bert}  &90.4 &79.3 &73.5 &90.2 &67.9   &80.2 \\
CLIP~\cite{radford2021learning}  &90.4 &79.8 &75.0 &90.2 &67.4   &80.6 \\
T5~\cite{raffel2020exploring}  &90.4 &80.0 &75.0 &\textbf{90.3} &67.9   &80.7 \\
MPNet~\cite{song2020mpnet}  &\textbf{90.4} &\textbf{80.7} &\textbf{76.8} &90.1 &\textbf{67.2}   &\textbf{81.0} \\
\hline
\end{tabular}  }
\caption{Results of different LLMs as text encoders.} 
\label{ablation_llm}
\end{table}

\textbf{Effects of Explicit Spatial alignment Module.}
The ESM aims to achieve coarse spatial alignment across modalities by integrating semantic relationships into the feature-level offset estimation. As shown in Table~\ref{ablation_model}, the introduction of the ESM elevates the mAP metric to $81.0\%$, with detection accuracy for each category improving to $90.4\%$, $80.6\%$, $76.3\%$, $90.1\%$, and $67.5\%$, respectively. Compared to using only the SAM and ISM, each evaluation metric demonstrates significant improvement. These experimental results and comparisons fully validate the effectiveness of our proposed Explicit Spatial alignment Module.

\textbf{Effects of different LLMs as Text Encoders.}
In this paper, we utilize ChatGPT to generate detailed text descriptions for each object category, and then employ a large language model to extract semantic features. Therefore, the capabilities of the selected large language model will directly impact the performance of our model. In the Table~\ref{ablation_llm}, we compare three commonly used large language models: BERT~\cite{kenton2019bert}, MPNet~\cite{song2020mpnet}, T5~\cite{raffel2020exploring}.In addition, we also include a multimodal large model, CLIP~\cite{radford2021learning}, in our comparison. Among them, MPNet performs the best, achieving an mAP score of $81.0\%$. Therefore, we use MPNet as the text encoder. 

\textbf{Effects of Different Text Contents.}
In this paper, we harness the powerful semantic representation capabilities of large language models by constructing text descriptions based on object categories. We then use a large language model as a text encoder to extract semantic features, guiding the model in progressively achieving semantic and spatial alignment. As shown in Table~\ref{ablation_text}, When we use category names directly as text input, the mAP score is $79.9\%$. Replacing category names with fine-grained object category descriptions generated with the assistance of ChatGPT increases the mAP score to $81.0\%$, reflecting an improvement of $1.1\%$ over the direct use of category names. This experimental result validates that fine-grained category descriptions can provide richer semantic information, facilitating better semantic and spatial alignment. 

\begin{table}[!htp]
\centering 
\resizebox{3.3in}{!}{
\begin{tabular}{c|cccccc}  
\hline 
\textbf{Method} &\textbf{Car}  &\textbf{Truck} &\textbf{Freight car} &\textbf{Bus}  &\textbf{Van} & \textbf{$mAP_{0.5}$} \\
\hline 
$w/o$ Text &90.3 &73.7 &68.7 &89.7 &59.7 &76.4 \\
Class Name &90.4 &77.5 &72.6 &90.3 &68.6   &79.9 \\
Class Description &90.4 &80.6 &76.3 &90.1 &67.5   &\textbf{81.0} \\
\hline
\end{tabular}  }
\caption{Results of different category text.} 
\label{ablation_text} 
\end{table}

\textbf{The Impact of text description quality on performance.}
To verify the impact of generated text quality on model performance, we compare two text perturbation strategies: random word replacement and random masking. The experimental results are shown in Table~\ref{text_per}. It is important to note that during text perturbation, the category words in the descriptions remain unchanged, and the masking and replacement ratio in these strategies is set to $50\%$. According to the results, when the quality of category descriptions degrades, the model performance decreases but still remains at a relatively high level. Additionally, we explore a fully automated text generation strategy, where ChatGPT automatically generates descriptive texts based on provided examples, which are then directly used for model training. Notably, this method results in only a $0.3\%$ performance drop compared to manual intervention. These experiments demonstrate that our approach can be effectively applied to other datasets.

\begin{table}[!htp]
\centering 
\resizebox{3.3in}{!}{
\begin{tabular}{c|cccccc}  
\hline 
\textbf{Method} &\textbf{Car}  &\textbf{Truck} &\textbf{Freight car} &\textbf{Bus}  &\textbf{Van} & \textbf{$mAP_{0.5}$} \\
\hline 
Random Mask &90.4 &79.7 &74.2 &90.3 &67.3   &80.4 \\
Random Replacement &90.4 &79.6 &74.3 &90.3 &67.6 &80.5 \\
Automatic Generation &90.4 &80.2 &75.7 &90.2 &67.8   &80.7 \\
Artificial Selection &90.4 &80.7 &76.8 &90.1 &67.2   &\textbf{81.0} \\
\hline
\end{tabular}  }
\caption{Results of different text perturbation strategies} 
\label{text_per} 
\end{table}

\subsection{Visualization} 

In this work, we visualize the detection results of the model, the similarity maps between the category description texts and RGB-IR images, and the feature maps after spatial-semantic alignment. As shown in Figure~\ref{detection_result}, our method is able to more accurately identify the object categories and locations. As shown in Figure~\ref{VT}, in our model, there is a high degree of responsiveness between the multimodal visual features of the object and the semantic features of the corresponding category. 
As shown in Figure~\ref{featuremap}, the features of our model are significantly enhanced, and there are clear boundaries around the object. Meanwhile, as indicated by the red dashed line area marked in the figure, our model can effectively perform spatial alignment between multimodal features. 


%% file: sec_arxiv/5_con.tex
\section{Conclusion}
\label{sec:con}

In this work, we propose LPANet, a multimodal UAV object detection method that is guided by a large language model. Our method leverages semantic features extracted by the large language model to progressively align both semantic and spatial features across modalities. Specifically, we first employ ChatGPT to generate fine-grained text descriptions for each object category and extract semantic features using the large language model. Next, we bring the semantic features of each object and multimodal visual features closer in a shared space to reduce semantic discrepancies across modalities. Subsequently, we integrate semantic relationships into the prediction of feature-level offsets, achieving coarse-grained spatial alignment across modalities. Finally, we aggregate key features from adjacent regions via cross-modal correlation to achieve implicit spatial alignment across modalities. Experimental results on the UAV object detection dataset demonstrate that our method outperforms current state-of-the-art approaches.